\documentclass[10pt, a4paper]{article}

\usepackage{lrec-coling2024} % this is the new style
\usepackage{booktabs}
\usepackage{amsfonts}
\usepackage{comment}
\usepackage{tikz}
\usetikzlibrary{shapes.geometric, arrows, positioning, calc}

\usepackage{amsmath}
\DeclareMathOperator*{\argmax}{arg\,max}

\DeclareMathOperator*{\Proba}{\mathbb{P}}

\def\promptedX{x'}

\def\Vocab{V}

\usepackage{xcolor} % add this line to use \textcolor
\definecolor{myred}{HTML}{FF3D3D}
\definecolor{myorange}{HTML}{FFAC6F}
\newcommand{\textred}[1]{\textcolor{myred}{\textbf{#1}}} % fix parameter errors
\newcommand{\textorange}[1]{\textcolor{myorange}{\textbf{#1}}} % fix parameter errors

\newcommand{\NumberOfModels}{72} % fix parameter errors
\newcommand{\NumberOfDatasets}{15} % fix parameter errors
\newcommand{\txtgitlab}{https://gitlab.lisn.upsaclay.fr/phd-pierre-lepagnol/classif_generation}
\title{Small Language Models are Good Too: An Empirical Study of Zero-Shot Classification}

\name{\normalsize Pierre Lepagnol$^{1,2}$, Thomas Gerald$^{1}$, Sahar Ghannay$^{1}$, Christophe Servan$^{1,3}$, Sophie Rosset$^1$} 

 \address{$^1$Université Paris-Saclay, CNRS, LISN, $^2$SCIAM, $^3$QWANT\\
         \{firstname.lastname\}@lisn.upsaclay.fr}

  \abstract{
    This study is part of the debate on the efficiency of large versus small language models for text classification by prompting.
    We assess the performance of small language models in zero-shot text classification, challenging the prevailing dominance of large models.
    Across 15 datasets, our investigation benchmarks language models from 77M to 40B parameters using different architectures and scoring functions. Our findings reveal that small models can effectively classify texts, getting on par with or surpassing their larger counterparts. We developed and shared a comprehensive open-source repository that encapsulates our methodologies. This research underscores the notion that bigger isn't always better, suggesting that resource-efficient small models may offer viable solutions for specific data classification challenges.
    \\ \newline  
    }

\begin{document}

\maketitleabstract
\section{Introduction}

Large Language Models (LLMs) have been massively favored over smaller models to solve tasks through prompting~\citep{Brown2020, Hoffmann2022, OpenAI2023, Chowdhery2022} in a zero-shot setting. 
However, while their utility is extensive, they come with challenges - they are resource-intensive, costly to employ, and their performances are not always warranted for every task~\citep{Nityasya2021}. 
Bigger models~\citep{Kaplan2020, Hoffmann2022} were built, always sophisticated datasets were necessary~\citep{Zhang2023} to achieve claimed performances.
Their perceived superior performance has typically made them the go-to choice for various tasks, even basic classification problems.

An application of LLMs is the generation of pseudo-labels through zero-shot prompting, a method often employed to construct labeled datasets~\citep{Smith2022}. 
As the field advances, we must ask: are large language models essential for effective data classification?

This study examines how well small models can match big models in creating labels using different datasets.
We want to see how small models perform in this zero-shot text classification and determine what makes them do well with specific data.
We are comparing how small and big models work with zero-shot prompting on various data sets to understand if we can get good results with less resources.

We believe this research is the beginning of understanding the true capabilities of LLMs when prompted for zero-shot classification tasks.

\paragraph{Our main contributions are:}

\begin{enumerate}
  \item We benchmark a large scale of language models (up to 70b parameters) fine-tuned on instructions-following datasets,
  with different architectures(encoder-decoder or decoder only) and sizes on many datasets in a zero-shot setting.
  \item We provide relatively strong evidence of the effectiveness of small models in zero-shot classification, and we show that the performances of small models are comparable to those of large models on many datasets in classification problems.
  \item We present a fully open-source repository encapsulating our proposed methodologies, thereby contributing to the integrity and robustness of research in this field.
  The code is available online in \href{\txtgitlab}{this repository}.
\end{enumerate}

The paper is organized as follows: Section~\ref{sec:rel_work} provides a literature review on related zero-shot approaches. 
Then, in Section~\ref{sec:method}, we describe the methodology we follow for this study.
Section~\ref{sec:results} presents the consequences given the different analyses.
Finally, we conclude in Section~\ref {sec:concl} and discuss future work and research directions.

\section{Related Work}\label{sec:rel_work}
The domain of zero-shot classification has previously been explored. These studies offer valuable insights and set the stage for our investigations.

\subsection{Zero-Shot Text Classification \& Prompting}
General zero-shot text classification aims to categorize texts into classes not part of the training dataset. 
It has caught the attention of many researchers because it removed the need for extra fine-tuning steps and labeled datasets. 
To effectively transfer knowledge from seen classes to unseen ones, there's a need for precise and distinguishing class descriptions, as noted by \citet{Xia2018} and \citet{Liu2019}.
Yet, these approaches depend on supervised data from recognized labels, which renders them unsuitable when there's a complete absence of labeled data for any given category.

\citet{Fei2022} enhances zero-shot classification by segmenting input texts and leveraging class-specific prompts. 
While \citet{Meng2020} proposed a strategy that employs label names combined with self-training tailored for zero-shot classification.
Many methods necessitate an unlabeled dataset or a knowledge base to extract pertinent topic words and facilitate self-training.
More recently,~\citet{Zhao2023} proposed to use k-Nearest-Neighbor on embeddings similarity to augment their verbalizers.
~\citet{Lu2023} proposed Perplexity Selection to select the best prompts in a zero-shot setting.  

\subsection*{Discussion}

While previous work focused on new methods to make language models better zero-shot learners, we want insight into model features and how well they perform.

% Moi je fais quoi par rapport à l'état de l'art.
% En quoi c'est différents et novateur.

\section{Experimental Setup}\label{sec:method}

Although authors of LLMs have compared their different model sizes\citep{Kaplan2020, Hoffmann2022}, this study widens this analysis by directly comparing different architectures on an extensive set of datasets.
We prompt various language models using 4 different scoring functions (see Section~\ref{subsec:scoring_func}) to classify sentences and report accuracy and F1 scores for each triple model-datasets-scoring function.

\subsection{Tasks \& Datasets}\label{subsec:datasets}

We examine a diverse set of \NumberOfDatasets{} datasets, curated to represent a broad spectrum of classification challenges.
We draw from datasets like \emph{AGNews}, with its 4 distinct classes, and \emph{BBCNews}, offering 5 unique categories for topic classification.
Sentiment classification is represented through binary choices like in \emph{ethos}~\citeplanguageresource{Mollas2022} and more granular datasets like \emph{sst-5}~\citeplanguageresource{Socher2013}.
Standard Spam classification tasks such as \emph{youtube} comments~\citeplanguageresource{Alberto2015} or \emph{sms}~\citeplanguageresource{misc_sms_spam_collection_228} are included.
Relation classification tasks are also included using datasets like \emph{semeval}~\citeplanguageresource{hendrickx2010semeval}.

The balance ratios across our chosen datasets varied extensively, from the perfectly balanced \emph{imdb} to those displaying significant imbalances like \emph{chemprot}~\citeplanguageresource{Krallinger2017}.

The complete list will be given in the final version.

\subsection{Metrics}
\label{subsec:Metrics}
We distinguish datasets on whether they are balanced using the balance ratio \textit{i.e.} the ratio between the majority class and the minority class.
The accuracy (acc) is used to evaluate binary tasks and balanced datasets, while the macro f1 (f1) score is used for the other tasks.

\subsection{Models}
Our study assesses a total of \NumberOfModels{} unique models.
We select both encoder-decoder models (like \texttt{T5}~\citep{Raffel2020}, \texttt{mT0}~\citep{Muennighoff2023}, and Bart~\cite{Lewis2020}) and causal-decoder-only models (such as \texttt{Llama}~\citep{Touvron2023} and \texttt{Falcon}~\citep{Penedo2023}).
We opt for various sizes for the same models, ranging from 77 million to hundreds of 40 billion parameters.
We called small language models, models within the size range 77M to 3B parameters. These models are comparatively smaller, ranging from 13 to 156 times less in parameter count than our largest model, \texttt{Falcon 40B}\footnote{We do not test \texttt{Falcon 180B}, as it was not released during our experiments}.
Moreover, at the time our study was conducted, \texttt{TinyStories}~\citep{eldan2023tinystories} models, which are on an even smaller scale, starting at 1M parameters.

These models were chosen based on their prevalence in literature, reported efficacy on similar tasks, and the fact that instruction-tuned versions were available for some of them.

Instruction-tuning refers to the strategy for fine-tuning a language model on instruction datasets \citep{longpre2023flan}.

The complete list will is given in appendix~\ref{sec:AppendixModels}. 

\subsection{Prompts \& Scoring Functions}

This section sets our research's specific prompts and scoring functions.
We follow~\cite{Brown2020} to craft simple prompts while ensuring domain relevance.
Additionally, we explore various scoring functions, assessing their impact on our models' performance.

\subsubsection{Prompts}
\label{subsec:prompts}
Our experiments' prompts are hand-crafted and designed to be simple and straightforward.

Prompts are either translated from the code-based labeling functions provided by the WRENCH benchmark~\citeplanguageresource{Zhang2021} or created from scratch.
They are tailored for each task, \textit{e.g.} prompts for the \emph{healthcare} dataset are framed differently from those for the financial dataset to ensure domain relevance and to maximize model comprehension.

For example, for the dataset \emph{sms}, the prompt is

\begin{table}[h]
  \centering
    \renewcommand{\arraystretch}{1.4}
  \begin{tabular}{c}
    Prompt\\
    \texttt{Is the following message spam?}\\\texttt{Answer by yes or no.\textbackslash{n}"\{TEXT\}"}\\
    Verbalizer\\
    \texttt{\{1:\texttt{"yes"}, 0:\texttt{"no"}\}} \\
  \end{tabular}
\end{table}

For \emph{bbcnews}, the prompt is
\begin{table}[h]
  \centering
    \renewcommand{\arraystretch}{1.4}
  \begin{tabular}{c}
    Prompt\\
    \texttt{""\{TEXT\}" is about "}\\
    Verbalizer\\
     \texttt{\{"0": "tech", "1": "business",}\\ \texttt{"2": "sport", "3": "entertainment",}\\ \texttt{"4": "politics"\}}.\\
  \end{tabular}
\end{table}

\subsubsection{Scoring Functions}\label{subsec:scoring_func}
In prompt-based classification, using a verbalizer mapping tokens to class labels is crucial for accurate classification.
As suggested by~\citep{Holtzman2022}, many valid sequences can represent the same concept, called \textit{surface form competition}.
For example, \texttt{"+"}, \texttt{"positive"}, \texttt{"More positive than the opposite"} could be used to represent the same concept of positivity for the sentiment analysis task.
As this competition exists, how verbalizers are designed could either mitigate or exacerbate the effects of surface form competition, thereby influencing the overall effectiveness of the prompt-based classification approach.
\citet{Zhao2023} uses k-Nearest-Neighbor for verbalizer construction and augments their verbalizers based on embeddings similarity.

We use several scoring functions to evaluate the impact of scoring functions on the performances of our models.
We describe in plain english these scoring function in appendix~\ref{sec:AppendixPrompts}.

\begin{table}[h!]
  \centering
  \renewcommand{\arraystretch}{1.8}
  \begin{tabular}{lr}
    Probability  &$\argmax_{i} \Proba(y_i|\promptedX)$ \\
    DCPMI &  $\argmax_{i} \frac{\Proba(y_i|\promptedX)}{\Proba(y_i|x_{\text{domain\_conditional}})}$\\
    PMI & $\argmax_{i} \frac{\Proba(y_i|\promptedX)}{\Proba(y_i|x_{\text{domain\_unconditional}})}$\\
    Similarity & $\argmax_{c_i \in C} \cos(e(t_0), e(y_i))$\footnote{$t_0=\argmax_{t \in \Vocab} \Proba(t|\promptedX)$}\\
  \end{tabular}
  \caption{Scoring functions from \citep{Holtzman2022}}
  \label{tab:scoring_methods}
\end{table}

\begin{table*}[h]

\resizebox{\textwidth}{!}{%
\begin{tabular}{lcccrr}
\toprule
dataset & SOTA Scores & Majority Class - Scores & Best Score & Model Used & Number of parameters \\
\midrule
agnews & 0.625 & 0.266 & \textred{0.734} & MBZUAI/LaMini-GPT-124M & 163.0 Millions \\
bbcnews & NaN & 0.236 & 0.869 & bigscience/mt0-large & 1.2 Billions \\
cdr & NaN & 0.676 & 0.717 & bigscience/bloomz-3b & 3.6 Billions \\
chemprot & 0.172 & 0.049 & \textred{0.192} & bigscience/bloomz-3b & 3.6 Billions \\
ethos & 0.667 & 0.566 & 0.597 & bigscience/bloomz-1b1 & 1.5 Billions \\
financial\_phrasebank & 0.528 & 0.254 & \textred{0.744} & MBZUAI/LaMini-GPT-774M & 838.4 Millions \\
imdb & 0.718 & 0.500 & \textred{0.933} & MBZUAI/LaMini-Flan-T5-783M & 783.2 Millions \\
semeval & 0.435 & 0.054 & 0.270 & bigscience/mt0-xxl & 12.9 Billions \\
sms & 0.340 & 0.464 & \textred{0.699} & mosaicml/mpt-7b & 6.6 Billions \\
spouse & 0.630 & 0.479 & 0.521 & gpt2 & 163.0 Millions \\
sst-2 & 0.710 & 0.501 & \textred{0.956} & bigscience/bloomz-3b & 3.6 Billions \\
sst-5 & 0.598 & 0.286 & 0.485 & tiiuae/falcon-40b-instruct & 41.8 Billions \\
trec & NaN & 0.072 & 0.324 & mosaicml/mpt-7b-instruct & 6.6 Billions \\
yelp & 0.888 & 0.522 & \textred{0.977} & MBZUAI/LaMini-Flan-T5-783M & 783.2 Millions \\
youtube & 0.468 & 0.528 & \textred{0.716} & tiiuae/falcon-40b & 41.8 Billions \\
\bottomrule
\end{tabular}
}
\caption{Table illustrating the performance metrics across various datasets:\\Columns present (1) the dataset name, (2) the reported state-of-the-art (SOTA) scores, (3) scores obtained when predicting the majority class, (4) the highest achieved scores (highlighted in red), (5) the model architectures associated with these top scores, and (6) the number of parameters for each respective model. Note the presence of NaN entries, signifying datasets where SOTA benchmarks have not been established or found.}\label{tab:sota_scores} 
\end{table*}

\subsection{Comparison}
\label{subsec:comparisons}
% \remsg{ajute plus d'information dans la légende du tableau, par exemple les métrics par datatset ...  }
We compare our results with Majority Voting (i.e predicting the class of the majority class in the dataset) and state-of-the-art (SOTA) Zero-Shot Learning methods.
Table~\ref{tab:sota_scores} presents the SOTA scores for each dataset\footnote{We removed scores from the \texttt{mT0} model for some datasets (\emph{agnews}, \emph{imdb}, \emph{yelp},\emph{trec}) because these models were trained on those datasets.}.

% Qu'est ce qu'on a fait ? Mesuré ? Et a quoi chacunes des mesures correspond ?

\subsection{Tools for Statistical Analysis}\label{subsec:tools}
For our analysis, we make use of three main statistical tools, detailed below:

\begin{description}
\item[The Biweight Midcorrelation Coefficient] is a robust alternative to Pearson's correlation coefficient to quantify the strength of association between two samples. It is designed to be less sensitive to outliers than other correlation coefficients like Pearson's correlation.
\item[Analysis of Covariance - ANCOVA] combines the techniques of ANOVA and regression to evaluate whether the means of a dependent variable are equal across levels of a categorical independent variable while statistically controlling for the effects of other continuous variables (covariates).
\item[Kruskal-Wallis Test] is a non-parametric method to test whether samples originate from the same distribution. We used it as a non-parametric method, which does not assume a normal distribution of the residuals, unlike the analogous standard one-way analysis of variance.
\end{description}

\section{Results}\label{sec:results}
We compare the performance of the LLM models on several datasets, studying the correlation with the number of parameters, the impact of the architecture, and the type of training strategy (instruction or not). 
Then, for the two types of architectures (encoder-decoder \& decoder-only), we study the impact of the instruction-tuning and the different scoring functions to understand the discriminating factors on performance.

\subsection{Data-based Analysis}
%We began with a dataset-based analysis.
For the dataset-based analysis, we propose to study: 1) the relationship between the task performances and the model sizes (the number of parameters), 2)  the task performances and the architecture,  and 3) whether the model was fine-tuned on instruction datasets.
% \remsg{j'ai réecrit le paragraphe pour bien montré les trois points à améliorer}
\subsubsection{Model size doesn't really matter}\label{subsubsec:Modelsize_Datasets}
\begin{figure*}[ht]
    \centering
    \includegraphics[width=1.1\textwidth]{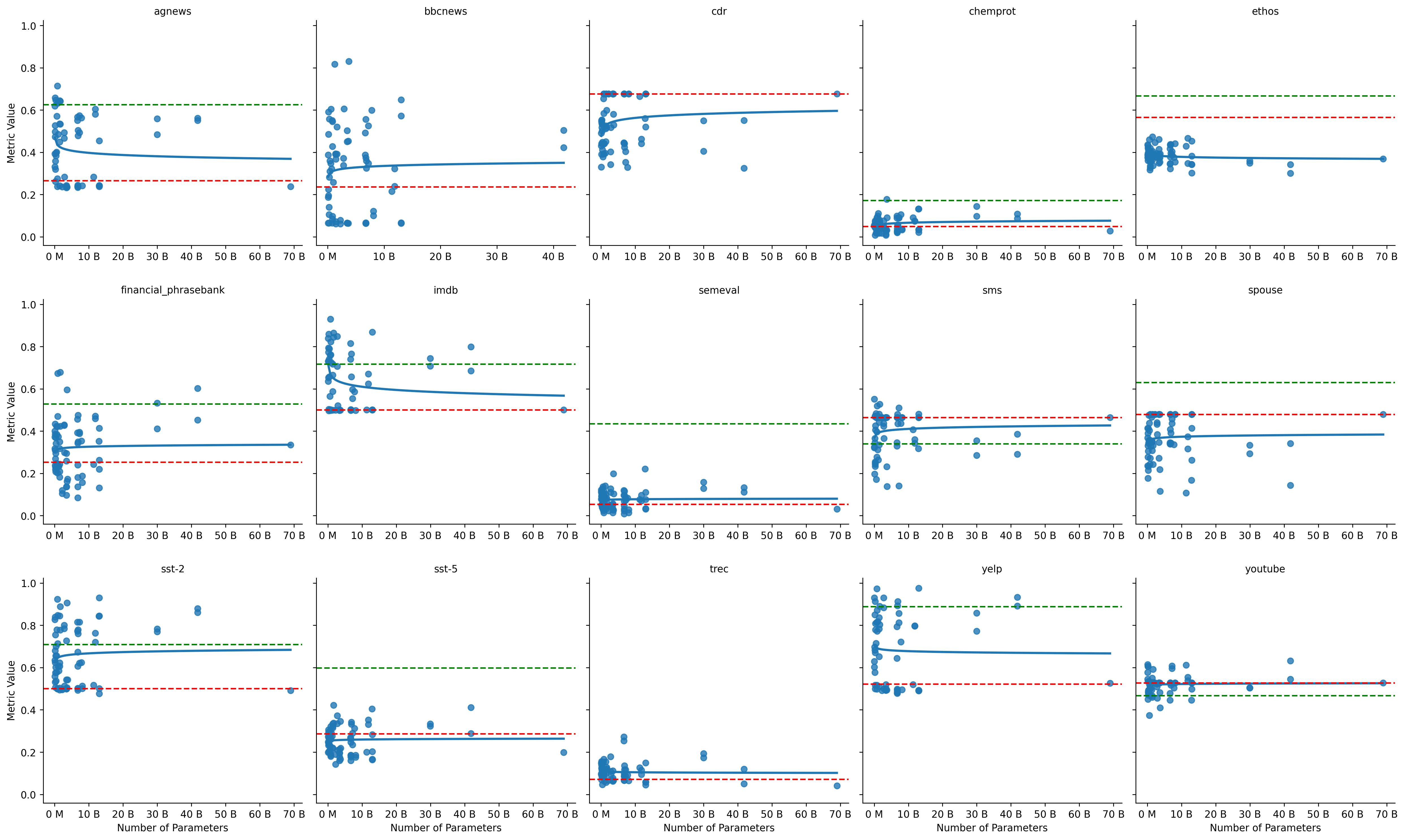}
    \caption{Performance Comparison of Different Model Sizes Across Datasets.}\label{fig:regplot_numparams_scores_per_datasets}
    \footnote{The figure uses a blue shaded region to indicate general trends, a red dashed line for scores from majority class predictions, and a green line to mark the current state-of-the-art (SOTA) scores for zero-shot prompting methodologies.}
\end{figure*}

% Figure + Description
Figure~\ref{fig:regplot_numparams_scores_per_datasets} presents the relationship between the number of parameters and the performance in terms of Acc/F1 scores across various datasets.
% \remsg{j'ai reformulé la phrase}
The correlations observed range from positive and negative to zero.

To further understand these correlations, we calculate the Biweight Midcorrelation Coefficient and associated p-values for each dataset. These findings are detailed in Table~\ref{tab:correlation_numparams_metric}.
\begin{table}[h]
    \resizebox{\columnwidth}{!}{
    \begin{tabular}{lcc}
        \toprule
        dataset & correlation coef & pvalue \\
        \midrule
        agnews & -0.1418 & \textorange{0.0536} \\
        bbcnews & 0.0489 & 0.4877 \\
        cdr & 0.2541 & \textred{0.0002} \\
        chemprot & 0.1318 & \textorange{0.0531} \\
        ethos & -0.1519 & \textred{0.0256} \\
        financial\_phrasebank & 0.0419 & 0.5406 \\
        imdb & -0.2862 & \textred{0.0001} \\
        semeval & -0.0506 & 0.4595 \\
        sms & -0.1209 & 0.0763 \\
        spouse & -0.0254 & 0.7106 \\
        sst-2 & 0.0755 & 0.2693 \\
        sst-5 & 0.0061 & 0.9293 \\
        trec & -0.1085 & 0.1403 \\
        yelp & -0.0620 & 0.4008 \\
        youtube & -0.0014 & 0.9836 \\
        \bottomrule
    \end{tabular}
    }
    \caption{The Biweight Midcorrelation Coefficients and P-values Indicating the Relationship Between Acc/F1 and Model Size (Log-Number of Parameters) Across Datasets}\label{tab:correlation_numparams_metric}
\end{table}

% interpretation 
From our analysis, 10 of 15 datasets show p-values exceeding 0.05, suggesting no significant link between Acc/F1 scores and model size. However, three datasets exhibit p-values below 0.05, indicating a notable correlation. Of these, the direction of correlation is positive for the \emph{cdr} dataset but negative for both \emph{ethos} and \emph{imdb} datasets. Two datasets, namely \emph{agnews} and \emph{chemprot}, present p-values near the 0.05 threshold, making their correlation inconclusive.

% Conclusion
In conclusion, while many datasets do not show a direct relationship between larger model sizes and improved performance, datasets like \emph{cdr}, \emph{ethos}, and \emph{imdb} do. Overall, the variance in the correlation coefficient across datasets suggests that model size isn't the sole determinant of performance.

\subsubsection{Impact of Architectural Choices on Performance}\label{subsubsec:Architecture_Datasets}

\begin{figure*}[h]
    \centering
    \includegraphics[width=1.\textwidth]{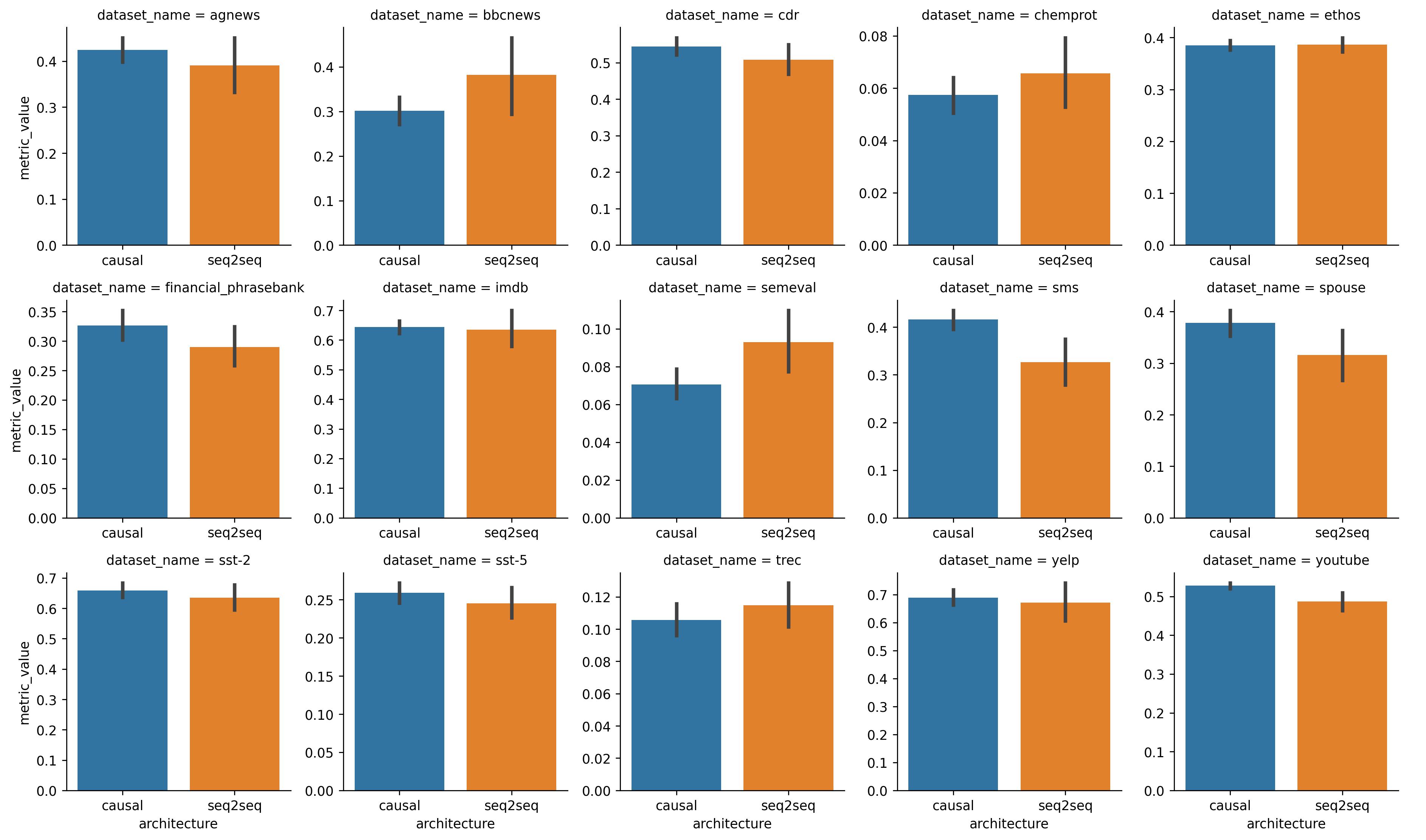}
    \caption{Performance Variation Across Different Architectures.}\label{fig:barplot_archtectures_scores_per_datasets}
\end{figure*}

% Figure + Description
Figure~\ref{fig:barplot_archtectures_scores_per_datasets} illustrates the performance variations between encoder-decoder and decoder-only architectures.

Using ANCOVA, we measure the impact of the architecture choice on Acc/F1 scores, while controlling the effect of the model size variable.
The results are presented in Table~\ref{tab:ancova_architecture_metric_per_datasets}.
\begin{table}%[h]
    \centering
    \resizebox{\columnwidth}{!}{
\begin{tabular}{lccr}
\toprule
Dataset & Statistic & Pvalue & Equal Variances \\
\midrule
agnews & 4.0676 & \textred{0.0452} & True \\
bbcnews & 7.0640 & \textred{0.0085} & False \\
cdr & 0.2519 & 0.6163 & True \\
chemprot & 4.4883 & \textred{0.0353} & True \\
ethos & 0.3945 & 0.5306 & False \\
financial\_phrasebank & 1.4592 & 0.2284 & False \\
imdb & 3.6687 & 0.0570 & True \\
semeval & 8.2301 & \textred{0.0045} & True \\
sms & 11.9951 & \textred{0.0006} & False \\
spouse & 4.7794 & \textred{0.0299} & True \\
sst-2 & 0.2501 & 0.6175 & True \\
sst-5 & 0.7852 & 0.3766 & True \\
trec & 0.3382 & 0.5616 & False \\
yelp & 0.7103 & 0.4004 & True \\
youtube & 18.0011 & \textred{0.0000} & False \\
\bottomrule
\end{tabular}
    }
\caption{ANCOVA Indicating the Impact of Architectures on Acc/F1 Across Datasets}\label{tab:ancova_architecture_metric_per_datasets}
\end{table}

% interpretation 
On one hand, 7 out of 15 datasets, namely \emph{agnews}, \emph{bbcnews}, \emph{chemprot}, \emph{semeval}, \emph{sms}, \emph{spouse}, and \emph{youtube}, show p-values bellow 0.05, suggesting there the architecture has a significant impact. 

On the other hand, datasets such as \emph{cdr}, \emph{ethos}, and \emph{financial\_phrasebank} remain unaffected by the architectural choice.
The \emph{imdb} dataset demonstrates a borderline significance.

In conclusion, while the model size might not be a dominant factor, the architectural choice significantly impacts performance across specific datasets.

\subsubsection{Impact of Instruction fine-tuning on performance}\label{subsubsec:Instruct_Datasets}
In the same way as architecture, we quantified the impact of instruction-tuning on performances while controlling the number of parameters.

\begin{figure*}[ht]
    \centering
    \includegraphics[width=1.0\textwidth]{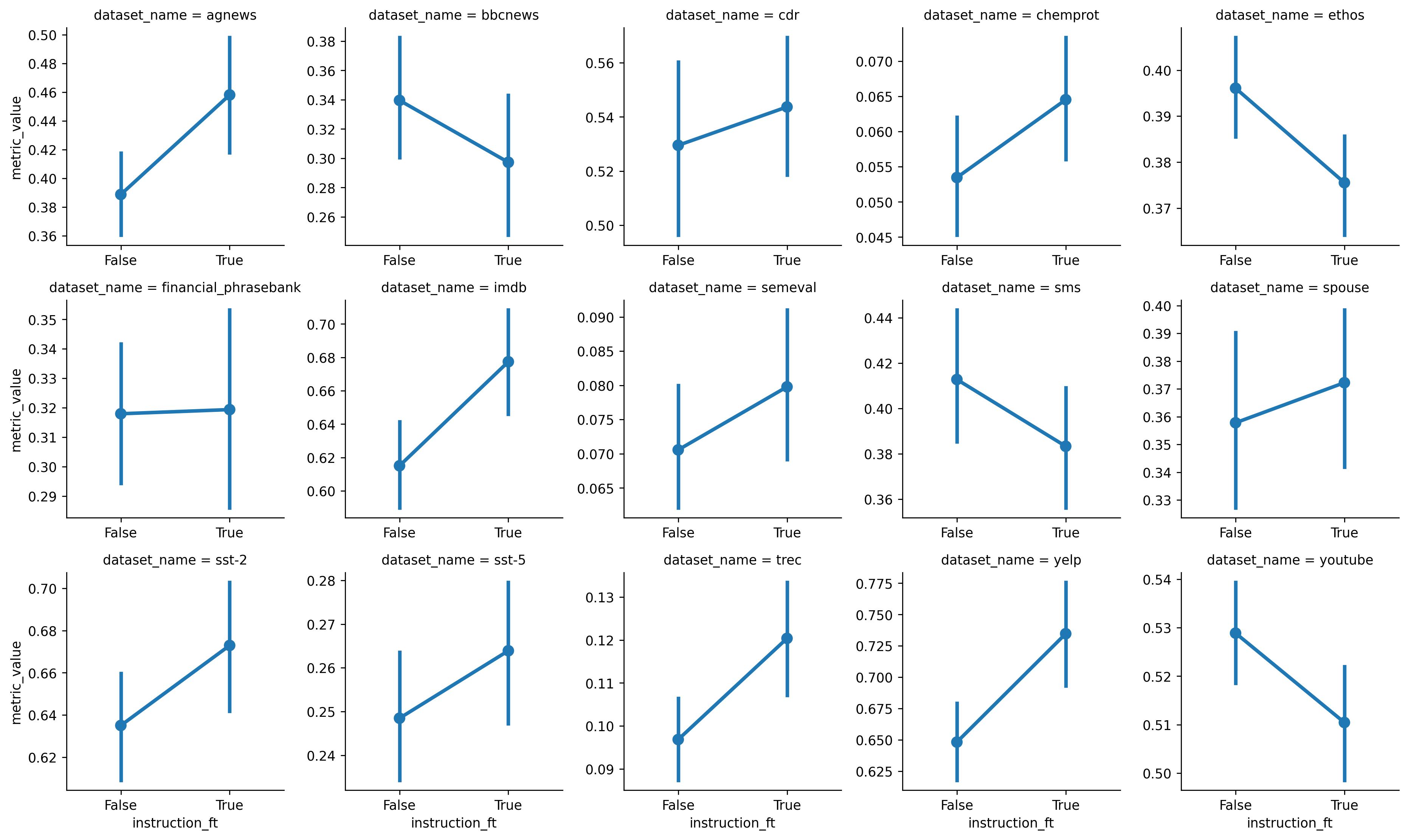}
    \caption{Performance Comparison between Instruction-Tuned models or not Across Datasets.}\label{fig:point_instruct_scores_per_datasets}
\end{figure*}

% Figure + Description
Figure~\ref{fig:point_instruct_scores_per_datasets} visually compares the impact of instruction-tuning and performance metrics (Acc/F1) across various datasets.

The y-axis of each graph displays the performance metric (Acc/F1).
The x-axis has two values: \texttt{False} and \texttt{True}, indicating whether instruction fine-tuning is applied to the model.

We use ANCOVA to test whether the means of our ACC/F1 scores are equal across modalities of instruction tuning while statistically controlling the effect of the number of parameters.

\begin{table}[h]
    \centering
    \resizebox{\columnwidth}{!}{
\begin{tabular}{lccr}
\toprule
dataset & statistic & pvalue & Equal Variances \\
\midrule
agnews & 10.5411 & \textred{0.0014} & True \\
bbcnews & 1.9492 & 0.1642 & True \\
cdr & 0.1635 & 0.6864 & True \\
chemprot & 2.3152 & 0.1296 & True \\
ethos & 5.8015 & \textred{0.0169} & True \\
financial\_phrasebank & 0.0001 & 0.9917 & False \\
imdb & 13.6945 & \textred{0.0003} & True \\
semeval & 1.4016 & 0.2378 & False \\
sms & 2.6667 & 0.1039 & True \\
spouse & 0.3379 & 0.5617 & True \\
sst-2 & 3.0055 & 0.0844 & False \\
sst-5 & 1.8271 & 0.1779 & True \\
trec & 8.3534 & \textred{0.0043} & False \\
yelp & 12.5571 & \textred{0.0005} & True \\
youtube & 5.8369 & \textred{0.0165} & True \\
\bottomrule
\end{tabular}
}
\caption{ANCOVA Indicating the Impact of The instruction Fine-Tunning on Acc/F1 Across Datasets}\label{tab:ancova_instruction_ft_metric_per_datasets}
\end{table}

% interpretation 
For many datasets, instruction fine-tuning improves performances when compared to not fine-tuning (\textit{e.g.}, \emph{agnews}, \emph{ethos}, \emph{imdb}, \emph{trec}, \emph{yelp}, and \emph{youtube}). This is evident from the graphical representation and the significant p-values from the ANCOVA.
Datasets like \emph{bbcnews}, \emph{youtube}, and \emph{sms} show a decrease in performance when instruction fine-tuning is applied, but ANCOVA tells us that it is not significant. While for \emph{ethos}, it is significant.

For other datasets, while there might be visual differences in performance with and without instruction fine-tuning, these differences aren't statistically significant based on the p-values.

% Conclusion
Therefore, while instruction fine-tuning has the potential to enhance model performance on many datasets, its impact may vary depending on the specific dataset.

\subsection{Architecture-based Analysis}
% \remsg{je ne comprend pas ma différence avec la section 4.1.2}
In our analysis, we shift our attention to which features among the model size, instruction-tuning, and scoring functions have an impact on performance.

\subsubsection{relationship between model size and performances per architecture}

% Figure + Description
Table~\ref{tab:correlation_numparams_metric_per_architecture} presents The Biweight Midcorrelation Coefficients between the model sizes (log-number of parameters) and performance metrics (Acc/F1) for either encoder-decoder and decoder-only.
\begin{table}[h]
    \centering
\begin{tabular}{lcc}
\toprule
dataset & correlation coef & pvalue \\
\midrule
causal & -0.0435 & \textred{0.0299} \\
seq2seq & 0.0065 & 0.8728 \\
\bottomrule
\end{tabular}
\caption{The Biweight Midcorrelation Coefficients and P-values Indicating the Relationship Between Acc/F1 and Model Size (Log-Number of Parameters) Across Architectures}\label{tab:correlation_numparams_metric_per_architecture}
\end{table}

% interpretation 
Table~\ref{tab:correlation_numparams_metric_per_architecture} shows a slight but significant correlation for decoder models but largely insignificant for encoder-decoder ones.

% Conclusion
This suggests that decoder-only could be more sensitive to the number of parameters; too many parameters could harm performance.

\subsubsection{Impact of Instruction Fine-tuning and Performances per architecture}
\begin{figure}[h]
    \centering
    \includegraphics[width=0.8\linewidth]{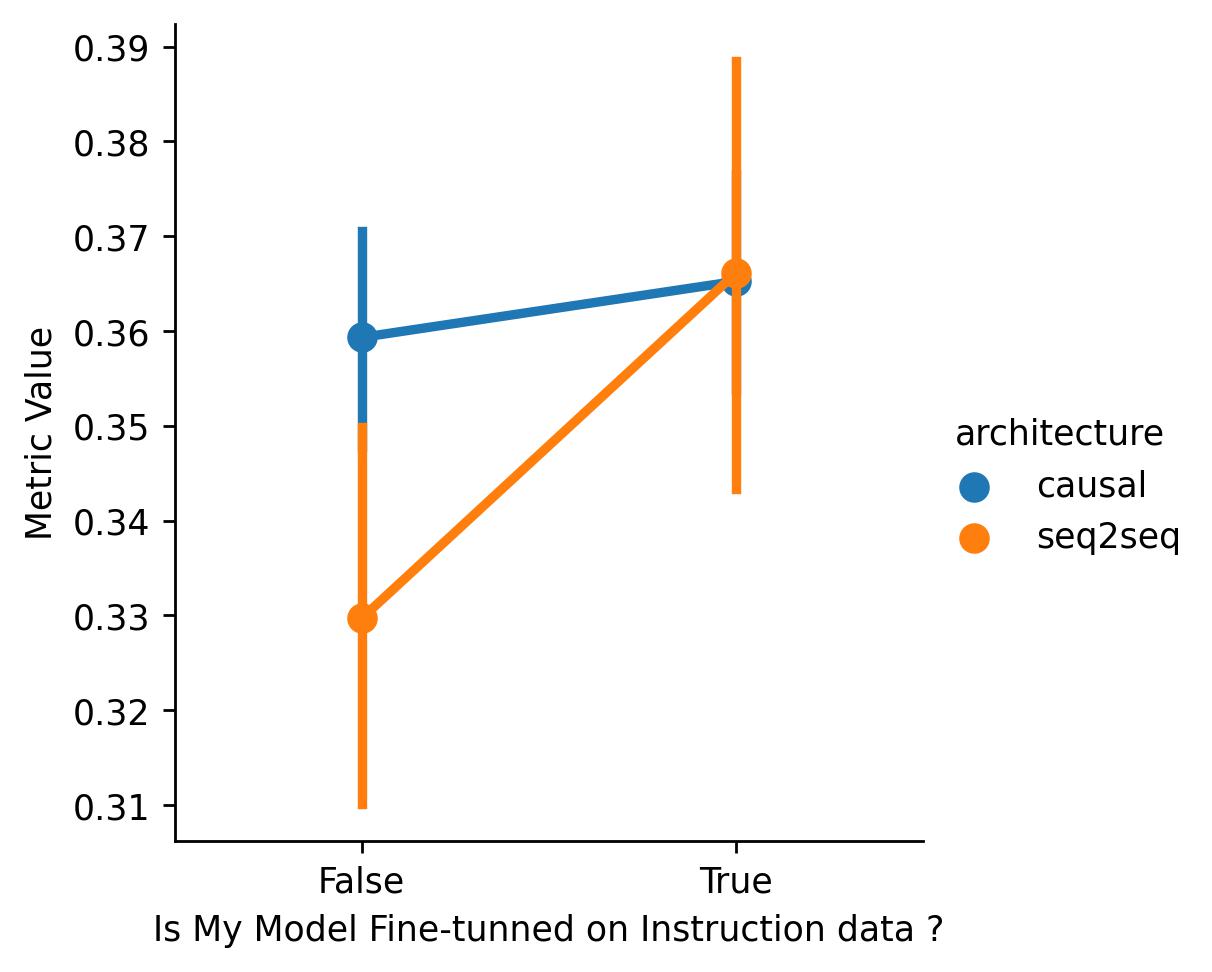}
    \caption{Performance Comparison between Instruction-Tuned models or not, Across Model Architecture}
    \label{fig:pointplot_modelsize_per_architecture}
\end{figure}

% Figure + Description
Figure~\ref{fig:pointplot_modelsize_per_architecture} visually compares the impact of instruction-tuning and performance metrics (Acc/F1) for the two architectures.

The y-axis is the performance metric (Acc/F1).
The x-axis has two values: \texttt{False} and \texttt{True}, indicating whether instruction fine-tuning is applied to the model.

An ANCOVA is made to quantify the impact of instruction-tuning on each architecture (encoder-decoder/decoder-only) while statistically controlling for the effect of the model size feature. Table~\ref{tab:ancova_instruction_ft_metric_per_architectures} reports statistics and p-values.

\begin{table}[h]
\begin{tabular}{lrrc}
\toprule
dataset & statistic & pvalue & Equal Variances \\
\midrule
causal & 0.1825 & 0.6693 & True \\
seq2seq & 6.9406 & \textred{0.0086} & False \\
\bottomrule
\end{tabular}
\caption{ANCOVA Indicating the Impact of instruction\_ft on Acc/F1 Across Architectures.}\label{tab:ancova_instruction_ft_metric_per_architectures}
\end{table}

% interpretation 
For the causal architecture, there is no significant impact of instruction-tuning on Acc/F1 scores. The p-value for the decoder-only architecture is 0.6693, much greater than 0.05.
For the seq2seq architecture, there is a significant impact of instruction tuning on Acc/F1 scores. The p-value for the encoder-decoder architecture is highlighted in red as 0.0086, less than 0.05.
Additionally, the variances between the groups for seq2seq are not equal.

% Conclusion
The difference in results between the two architectures suggests that the impact of instruction-tuning might be architecture-dependent.
Both the graphical analysis and the ANCOVA show an effect of instruction-tuning on encoder-decoder architecture.

\subsubsection{Impact of Scoring Functions and Performances per architecture}

% Figure + Description
Table~\ref{tab:scoring_functions_per_archi} reports the ANCOVA results of the impact of different scoring functions on performances for the two architectures.

\begin{table}[h]
    \centering
    \resizebox{\columnwidth}{!}{
\begin{tabular}{lccc}
\toprule
Architecture & statistic & pvalue & Equal Variances \\
\midrule
causal & 0.6711 & 0.5113 & False \\
seq2seq & 0.5003 & 0.6066 & True \\
\bottomrule
\end{tabular}
}
\caption{ANCOVA Indicating the Impact of Scoring Functions on Acc/F1 Across Architectures.}\label{tab:scoring_functions_per_archi}
\end{table}

% interpretation 
For both encoder-decoder and decoder-only models, values are above the standard 0.05 by a large margin. This suggests no significant impact on the choice of scoring functions.

% Conclusion
To sum it up, no matter which model architecture we look at, the choice of scoring function doesn't seem to affect more than another.

\section{Conclusion \& Perspectives}\label{sec:concl}
This paper aimed to understand better whether we need large models to tackle classification problems through prompting.

The performance of LLM models varies based on multiple factors, including model size, architectural choices, and fine-tuning strategies.
While larger model sizes do not consistently lead to improved performance across all datasets, the architectural choice significantly influences outcomes on specific datasets.
The impact of instruction fine-tuning is also evident, but its efficacy is dependent on the architecture.
Notably, the choice of scoring function doesn't seem to make a marked difference in performance.

A comprehensive study of other emerging architectures, such as RWKV architecture \citep{peng2023rwkv} or Retentive Network \citep{sun2023retentive}, could bring nuances and detail to this analysis.
The varied impact of instruction fine-tuning across datasets suggests the need for more advanced fine-tuning techniques like incorporating information retrieval during fine-tuning to ensure even better classification performances during zero-shot prompting.

\section{Limitations}
We limit this evaluation to simple prompting methods and hand-crafted, unoptimized prompts.
We also provide a single prompt for each dataset. 

We focused on causal-decoder-only and encoder-decoder models without comparing them with encoder-only or non-causal decoders as recently released models focused on those architectures.

We did not mention external factors such as pre-training time, data quality, or potential biases in the datasets.
These external factors might impact the results or the generalizability of the conclusions.

The choice and assumptions of the statistical tools could influence the results.
There might be newer or specialized models not included in this study, which could exhibit different behaviors.

\section{Acknowledgements}

This work is supported by the ANRT (Association nationale de la recherche et de la technologie) with a CIFRE fellowship granted to SCIAM\footnote{\href{https://www.sciam.fr}{https://www.sciam.fr/}}. (CIFRE N°2022/1608)

This work was performed using HPC resources from GENCI–IDRIS (Grant 2023-AD011014242).

\section*{Ethics Statement}
It is worth noting that the behavior of our downstream models is subject to biases inherited
from the dataset it was trained, as no alignment nor specific filtering was done.
We envision the same research progress in reducing anti-social behaviors in LLMs can also be applied to improve smaller language models.

%\nocite{*}
\section{Bibliographical References}\label{sec:reference}

\bibliographystyle{lrec-coling2024-natbib}
\bibliography{dataset_complexity, intro}

\begin{thebibliography}{7}
\expandafter\ifx\csname natexlab\endcsname\relax\def\natexlab#1{#1}\fi

\bibitem[{Alberto et~al.(2015)Alberto, Lochter, and Almeida}]{Alberto2015}
Tulio~C. Alberto, Johannes~V. Lochter, and Tiago~A. Almeida. 2015.
\newblock \href {https://doi.org/10.1109/ICMLA.2015.37} {{TubeSpam}: {Comment}
  {Spam} {Filtering} on {YouTube}}.
\newblock \emph{2015 IEEE 14th International Conference on Machine Learning and
  Applications (ICMLA)}, pages 138--143.

\bibitem[{Almeida and Hidalgo(2012)}]{misc_sms_spam_collection_228}
Tiago Almeida and Jos Hidalgo. 2012.
\newblock {SMS Spam Collection}.
\newblock UCI Machine Learning Repository.
\newblock {DOI}: https://doi.org/10.24432/C5CC84.

\bibitem[{Hendrickx et~al.(2010)Hendrickx, Kim, Kozareva, Nakov,
  {\'O}~S{\'e}aghdha, Pad{\'o}, Pennacchiotti, Romano, and
  Szpakowicz}]{hendrickx2010semeval}
Iris Hendrickx, Su~Nam Kim, Zornitsa Kozareva, Preslav Nakov, Diarmuid
  {\'O}~S{\'e}aghdha, Sebastian Pad{\'o}, Marco Pennacchiotti, Lorenza Romano,
  and Stan Szpakowicz. 2010.
\newblock \href {https://aclanthology.org/S10-1006} {{S}em{E}val-2010 task 8:
  Multi-way classification of semantic relations between pairs of nominals}.
\newblock In \emph{Proceedings of the 5th International Workshop on Semantic
  Evaluation}, pages 33--38, Uppsala, Sweden. Association for Computational
  Linguistics.

\bibitem[{Krallinger et~al.(2017)Krallinger, Rabal, Akhondi, Pérez,
  Santamaría, Rodríguez, Tsatsaronis, Intxaurrondo, López, Nandal, Buel,
  Chandrasekhar, Rodenburg, Lægreid, Doornenbal, Oyarzábal, Lourenço, and
  Valencia}]{Krallinger2017}
Martin Krallinger, O.~Rabal, S.~Akhondi, M.~Pérez, J.~Santamaría, Gael~Pérez
  Rodríguez, G.~Tsatsaronis, Ander Intxaurrondo, José Antonio~Baso López,
  U.~Nandal, E.~V. Buel, A.~Chandrasekhar, Marleen Rodenburg, A.~Lægreid,
  Marius~A. Doornenbal, J.~Oyarzábal, A.~Lourenço, and A.~Valencia. 2017.
\newblock \href
  {https://www.semanticscholar.org/paper/Overview-of-the-BioCreative-VI-chemical-protein-Krallinger-Rabal/eed781f498b563df5a9e8a241c67d63dd1d92ad5}
  {Overview of the {BioCreative} {VI} chemical-protein interaction {Track}}.

\bibitem[{Mollas et~al.(2022)Mollas, Chrysopoulou, Karlos, and
  Tsoumakas}]{Mollas2022}
Ioannis Mollas, Zoe Chrysopoulou, Stamatis Karlos, and Grigorios Tsoumakas.
  2022.
\newblock \href {https://doi.org/10.1007/s40747-021-00608-2} {{ETHOS}: an
  {Online} {Hate} {Speech} {Detection} {Dataset}}.
\newblock \emph{Complex \& Intelligent Systems}, 8(6):4663--4678.
\newblock ArXiv:2006.08328 [cs, stat].

\bibitem[{Socher et~al.(2013)Socher, Perelygin, Wu, Chuang, Manning, Ng, and
  Potts}]{Socher2013}
Richard Socher, Alex Perelygin, Jean Wu, Jason Chuang, Christopher~D. Manning,
  Andrew Ng, and Christopher Potts. 2013.
\newblock \href {https://aclanthology.org/D13-1170} {Recursive {Deep} {Models}
  for {Semantic} {Compositionality} {Over} a {Sentiment} {Treebank}}.
\newblock In \emph{Proceedings of the 2013 {Conference} on {Empirical}
  {Methods} in {Natural} {Language} {Processing}}, pages 1631--1642, Seattle,
  Washington, USA. Association for Computational Linguistics.

\bibitem[{Zhang et~al.(2021)Zhang, Yu, Li, Wang, Yang, Yang, and
  Ratner}]{Zhang2021}
Jieyu Zhang, Yue Yu, Yinghao Li, Yujing Wang, Yaming Yang, Mao Yang, and
  Alexander~J. Ratner. 2021.
\newblock \href
  {https://www.semanticscholar.org/paper/WRENCH%3A-A-Comprehensive-Benchmark-for-Weak-Zhang-Yu/3ba529f732d3c4a31e9ce57f1c78ddf911846bf4}
  {{WRENCH}: {A} {Comprehensive} {Benchmark} for {Weak} {Supervision}}.
\newblock \emph{ArXiv}.

\end{thebibliography}


\begin{thebibliography}{27}
\expandafter\ifx\csname natexlab\endcsname\relax\def\natexlab#1{#1}\fi

\bibitem[{Biderman et~al.(2023)Biderman, Schoelkopf, Anthony, Bradley, O'Brien,
  Hallahan, Khan, Purohit, Prashanth, Raff, Skowron, Sutawika, and van~der
  Wal}]{Biderman2023}
Stella Biderman, Hailey Schoelkopf, Quentin Anthony, Herbie Bradley, Kyle
  O'Brien, Eric Hallahan, Mohammad~Aflah Khan, Shivanshu Purohit, USVSN~Sai
  Prashanth, Edward Raff, Aviya Skowron, Lintang Sutawika, and Oskar van~der
  Wal. 2023.
\newblock \href {https://doi.org/10.48550/arXiv.2304.01373} {Pythia: {A}
  {Suite} for {Analyzing} {Large} {Language} {Models} {Across} {Training} and
  {Scaling}}.
\newblock Technical report.
\newblock ArXiv:2304.01373 [cs] type: article.

\bibitem[{Brown et~al.(2020)Brown, Mann, Ryder, Subbiah, Kaplan, Dhariwal,
  Neelakantan, Shyam, Sastry, Askell, Agarwal, Herbert-Voss, Krueger, Henighan,
  Child, Ramesh, Ziegler, Wu, Winter, Hesse, Chen, Sigler, Litwin, Gray, Chess,
  Clark, Berner, McCandlish, Radford, Sutskever, and Amodei}]{Brown2020}
Tom Brown, Benjamin Mann, Nick Ryder, Melanie Subbiah, Jared~D Kaplan, Prafulla
  Dhariwal, Arvind Neelakantan, Pranav Shyam, Girish Sastry, Amanda Askell,
  Sandhini Agarwal, Ariel Herbert-Voss, Gretchen Krueger, Tom Henighan, Rewon
  Child, Aditya Ramesh, Daniel Ziegler, Jeffrey Wu, Clemens Winter, Chris
  Hesse, Mark Chen, Eric Sigler, Mateusz Litwin, Scott Gray, Benjamin Chess,
  Jack Clark, Christopher Berner, Sam McCandlish, Alec Radford, Ilya Sutskever,
  and Dario Amodei. 2020.
\newblock \href
  {https://proceedings.neurips.cc/paper_files/paper/2020/hash/1457c0d6bfcb4967418bfb8ac142f64a-Abstract.html}
  {Language {Models} are {Few}-{Shot} {Learners}}.
\newblock In \emph{Advances in {Neural} {Information} {Processing} {Systems}},
  volume~33, pages 1877--1901. Curran Associates, Inc.

\bibitem[{Chowdhery et~al.(2022)Chowdhery, Narang, Devlin, Bosma, Mishra,
  Roberts, Barham, Chung, Sutton, Gehrmann, Schuh, Shi, Tsvyashchenko, Maynez,
  Rao, Barnes, Tay, Shazeer, Prabhakaran, Reif, Du, Hutchinson, Pope, Bradbury,
  Austin, Isard, Gur-Ari, Yin, Duke, Levskaya, Ghemawat, Dev, Michalewski,
  Garcia, Misra, Robinson, Fedus, Zhou, Ippolito, Luan, Lim, Zoph, Spiridonov,
  Sepassi, Dohan, Agrawal, Omernick, Dai, Pillai, Pellat, Lewkowycz, Moreira,
  Child, Polozov, Lee, Zhou, Wang, Saeta, Diaz, Firat, Catasta, Wei,
  Meier-Hellstern, Eck, Dean, Petrov, and Fiedel}]{Chowdhery2022}
Aakanksha Chowdhery, Sharan Narang, Jacob Devlin, Maarten Bosma, Gaurav Mishra,
  Adam Roberts, Paul Barham, Hyung~Won Chung, Charles Sutton, Sebastian
  Gehrmann, Parker Schuh, Kensen Shi, Sasha Tsvyashchenko, Joshua Maynez,
  Abhishek Rao, Parker Barnes, Yi~Tay, Noam Shazeer, Vinodkumar Prabhakaran,
  Emily Reif, Nan Du, Ben Hutchinson, Reiner Pope, James Bradbury, Jacob
  Austin, Michael Isard, Guy Gur-Ari, Pengcheng Yin, Toju Duke, Anselm
  Levskaya, Sanjay Ghemawat, Sunipa Dev, Henryk Michalewski, Xavier Garcia,
  Vedant Misra, Kevin Robinson, Liam Fedus, Denny Zhou, Daphne Ippolito, David
  Luan, Hyeontaek Lim, Barret Zoph, Alexander Spiridonov, Ryan Sepassi, David
  Dohan, Shivani Agrawal, Mark Omernick, Andrew~M. Dai,
  Thanumalayan~Sankaranarayana Pillai, Marie Pellat, Aitor Lewkowycz, Erica
  Moreira, Rewon Child, Oleksandr Polozov, Katherine Lee, Zongwei Zhou, Xuezhi
  Wang, Brennan Saeta, Mark Diaz, Orhan Firat, Michele Catasta, Jason Wei,
  Kathy Meier-Hellstern, Douglas Eck, Jeff Dean, Slav Petrov, and Noah Fiedel.
  2022.
\newblock \href {http://arxiv.org/abs/2204.02311} {{PaLM}: {Scaling} {Language}
  {Modeling} with {Pathways}}.
\newblock Technical report.
\newblock ArXiv:2204.02311 [cs] type: article.

\bibitem[{Eldan and Li(2023)}]{eldan2023tinystories}
Ronen Eldan and Yuanzhi Li. 2023.
\newblock \href {http://arxiv.org/abs/2305.07759} {Tinystories: How small can
  language models be and still speak coherent english?}

\bibitem[{Fei et~al.(2022)Fei, Meng, Nie, Wattenhofer, and Sachan}]{Fei2022}
Yu~Fei, Zhao Meng, Ping Nie, Roger Wattenhofer, and Mrinmaya Sachan. 2022.
\newblock \href {https://doi.org/10.18653/v1/2022.emnlp-main.587} {Beyond
  prompting: {Making} {Pre}-trained {Language} {Models} {Better} {Zero}-shot
  {Learners} by {Clustering} {Representations}}.
\newblock In \emph{Proceedings of the 2022 {Conference} on {Empirical}
  {Methods} in {Natural} {Language} {Processing}}, pages 8560--8579, Abu Dhabi,
  United Arab Emirates. Association for Computational Linguistics.

\bibitem[{Hoffmann et~al.(2022)Hoffmann, Borgeaud, Mensch, Buchatskaya, Cai,
  Rutherford, Casas, Hendricks, Welbl, Clark, Hennigan, Noland, Millican,
  Driessche, Damoc, Guy, Osindero, Simonyan, Elsen, Rae, Vinyals, and
  Sifre}]{Hoffmann2022}
Jordan Hoffmann, Sebastian Borgeaud, Arthur Mensch, Elena Buchatskaya, Trevor
  Cai, Eliza Rutherford, Diego de~Las Casas, Lisa~Anne Hendricks, Johannes
  Welbl, Aidan Clark, Tom Hennigan, Eric Noland, Katie Millican, George van~den
  Driessche, Bogdan Damoc, Aurelia Guy, Simon Osindero, Karen Simonyan, Erich
  Elsen, Jack~W. Rae, Oriol Vinyals, and Laurent Sifre. 2022.
\newblock \href {http://arxiv.org/abs/2203.15556} {Training {Compute}-{Optimal}
  {Large} {Language} {Models}}.
\newblock Technical report.
\newblock ArXiv:2203.15556 [cs] type: article.

\bibitem[{Holtzman et~al.(2022)Holtzman, West, Shwartz, Choi, and
  Zettlemoyer}]{Holtzman2022}
Ari Holtzman, Peter West, Vered Shwartz, Yejin Choi, and Luke Zettlemoyer.
  2022.
\newblock \href {https://doi.org/10.48550/arXiv.2104.08315} {Surface {Form}
  {Competition}: {Why} the {Highest} {Probability} {Answer} {Isn}'t {Always}
  {Right}}.
\newblock Technical report.
\newblock ArXiv:2104.08315 [cs] type: article.

\bibitem[{Kaplan et~al.(2020)Kaplan, McCandlish, Henighan, Brown, Chess, Child,
  Gray, Radford, Wu, and Amodei}]{Kaplan2020}
Jared Kaplan, Sam McCandlish, Tom Henighan, Tom~B. Brown, Benjamin Chess, Rewon
  Child, Scott Gray, Alec Radford, Jeffrey Wu, and Dario Amodei. 2020.
\newblock \href {https://doi.org/10.48550/arXiv.2001.08361} {Scaling {Laws} for
  {Neural} {Language} {Models}}.
\newblock Technical report.
\newblock ArXiv:2001.08361 [cs, stat] type: article.

\bibitem[{Lewis et~al.(2020)Lewis, Liu, Goyal, Ghazvininejad, Mohamed, Levy,
  Stoyanov, and Zettlemoyer}]{Lewis2020}
Mike Lewis, Yinhan Liu, Naman Goyal, Marjan Ghazvininejad, Abdelrahman Mohamed,
  Omer Levy, Veselin Stoyanov, and Luke Zettlemoyer. 2020.
\newblock \href {https://doi.org/10.18653/v1/2020.acl-main.703} {{BART}:
  {Denoising} {Sequence}-to-{Sequence} {Pre}-training for {Natural} {Language}
  {Generation}, {Translation}, and {Comprehension}}.
\newblock In \emph{Proceedings of the 58th {Annual} {Meeting} of the
  {Association} for {Computational} {Linguistics}}, pages 7871--7880, Online.
  Association for Computational Linguistics.

\bibitem[{Liu et~al.(2019)Liu, Zhang, Fan, Fu, Li, Wu, and Lam}]{Liu2019}
Han Liu, Xiaotong Zhang, Lu~Fan, Xuandi Fu, Qimai Li, Xiao-Ming Wu, and
  Albert~Y.S. Lam. 2019.
\newblock \href {https://doi.org/10.18653/v1/D19-1486} {Reconstructing
  {Capsule} {Networks} for {Zero}-shot {Intent} {Classification}}.
\newblock In \emph{Proceedings of the 2019 {Conference} on {Empirical}
  {Methods} in {Natural} {Language} {Processing} and the 9th {International}
  {Joint} {Conference} on {Natural} {Language} {Processing}
  ({EMNLP}-{IJCNLP})}, pages 4799--4809, Hong Kong, China. Association for
  Computational Linguistics.

\bibitem[{Longpre et~al.(2023)Longpre, Hou, Vu, Webson, Chung, Tay, Zhou, Le,
  Zoph, Wei, and Roberts}]{longpre2023flan}
Shayne Longpre, Le~Hou, Tu~Vu, Albert Webson, Hyung~Won Chung, Yi~Tay, Denny
  Zhou, Quoc~V. Le, Barret Zoph, Jason Wei, and Adam Roberts. 2023.
\newblock \href {http://arxiv.org/abs/2301.13688} {The flan collection:
  Designing data and methods for effective instruction tuning}.

\bibitem[{Lu et~al.(2023)Lu, Zhu, Han, Zhao, Mac~Namee, and Tan}]{Lu2023}
Jinghui Lu, Dongsheng Zhu, Weidong Han, Rui Zhao, Brian Mac~Namee, and Fei Tan.
  2023.
\newblock \href {https://doi.org/10.18653/v1/2023.acl-long.128} {What makes
  pre-trained language models better zero-shot learners?}
\newblock In \emph{Proceedings of the 61st Annual Meeting of the Association
  for Computational Linguistics (Volume 1: Long Papers)}, pages 2288--2303,
  Toronto, Canada. Association for Computational Linguistics.

\bibitem[{Meng et~al.(2020)Meng, Zhang, Huang, Xiong, Ji, Zhang, and
  Han}]{Meng2020}
Yu~Meng, Yunyi Zhang, Jiaxin Huang, Chenyan Xiong, Heng Ji, Chao Zhang, and
  Jiawei Han. 2020.
\newblock \href {https://doi.org/10.18653/v1/2020.emnlp-main.724} {Text
  {Classification} {Using} {Label} {Names} {Only}: {A} {Language} {Model}
  {Self}-{Training} {Approach}}.
\newblock In \emph{Proceedings of the 2020 {Conference} on {Empirical}
  {Methods} in {Natural} {Language} {Processing} ({EMNLP})}, pages 9006--9017,
  Online. Association for Computational Linguistics.

\bibitem[{Muennighoff et~al.(2023)Muennighoff, Wang, Sutawika, Roberts,
  Biderman, Scao, Bari, Shen, Yong, Schoelkopf, Tang, Radev, Aji, Almubarak,
  Albanie, Alyafeai, Webson, Raff, and Raffel}]{Muennighoff2023}
Niklas Muennighoff, Thomas Wang, Lintang Sutawika, Adam Roberts, Stella
  Biderman, Teven~Le Scao, M.~Saiful Bari, Sheng Shen, Zheng-Xin Yong, Hailey
  Schoelkopf, Xiangru Tang, Dragomir Radev, Alham~Fikri Aji, Khalid Almubarak,
  Samuel Albanie, Zaid Alyafeai, Albert Webson, Edward Raff, and Colin Raffel.
  2023.
\newblock \href {https://doi.org/10.48550/arXiv.2211.01786} {Crosslingual
  {Generalization} through {Multitask} {Finetuning}}.
\newblock Technical report.
\newblock ArXiv:2211.01786 [cs] type: article.

\bibitem[{Nityasya et~al.(2021)Nityasya, Wibowo, Prasojo, and
  Aji}]{Nityasya2021}
Made~Nindyatama Nityasya, Haryo~Akbarianto Wibowo, Radityo~Eko Prasojo, and
  Alham~Fikri Aji. 2021.
\newblock \href {http://arxiv.org/abs/2012.08958} {Costs to {Consider} in
  {Adopting} {NLP} for {Your} {Business}}.
\newblock Technical report.
\newblock ArXiv:2012.08958 [cs] type: article.

\bibitem[{OpenAI(2023)}]{OpenAI2023}
OpenAI. 2023.
\newblock \href {https://doi.org/10.48550/arXiv.2303.08774} {{GPT}-4
  {Technical} {Report}}.
\newblock Technical report.
\newblock ArXiv:2303.08774 [cs] type: article.

\bibitem[{Penedo et~al.(2023)Penedo, Malartic, Hesslow, Cojocaru, Cappelli,
  Alobeidli, Pannier, Almazrouei, and Launay}]{Penedo2023}
Guilherme Penedo, Quentin Malartic, Daniel Hesslow, Ruxandra Cojocaru,
  Alessandro Cappelli, Hamza Alobeidli, Baptiste Pannier, Ebtesam Almazrouei,
  and Julien Launay. 2023.
\newblock \href {https://doi.org/10.48550/arXiv.2306.01116} {The {RefinedWeb}
  {Dataset} for {Falcon} {LLM}: {Outperforming} {Curated} {Corpora} with {Web}
  {Data}, and {Web} {Data} {Only}}.
\newblock Technical report.
\newblock ArXiv:2306.01116 [cs] type: article.

\bibitem[{Peng et~al.(2023)Peng, Alcaide, Anthony, Albalak, Arcadinho, Cao,
  Cheng, Chung, Grella, GV, He, Hou, Kazienko, Kocon, Kong, Koptyra, Lau,
  Mantri, Mom, Saito, Tang, Wang, Wind, Wozniak, Zhang, Zhang, Zhao, Zhou, Zhu,
  and Zhu}]{peng2023rwkv}
Bo~Peng, Eric Alcaide, Quentin Anthony, Alon Albalak, Samuel Arcadinho, Huanqi
  Cao, Xin Cheng, Michael Chung, Matteo Grella, Kranthi~Kiran GV, Xuzheng He,
  Haowen Hou, Przemyslaw Kazienko, Jan Kocon, Jiaming Kong, Bartlomiej Koptyra,
  Hayden Lau, Krishna Sri~Ipsit Mantri, Ferdinand Mom, Atsushi Saito, Xiangru
  Tang, Bolun Wang, Johan~S. Wind, Stansilaw Wozniak, Ruichong Zhang, Zhenyuan
  Zhang, Qihang Zhao, Peng Zhou, Jian Zhu, and Rui-Jie Zhu. 2023.
\newblock \href {http://arxiv.org/abs/2305.13048} {Rwkv: Reinventing rnns for
  the transformer era}.

\bibitem[{Raffel et~al.(2020)Raffel, Shazeer, Roberts, Lee, Narang, Matena,
  Zhou, Li, and Liu}]{Raffel2020}
Colin Raffel, Noam Shazeer, Adam Roberts, Katherine Lee, Sharan Narang, Michael
  Matena, Yanqi Zhou, Wei Li, and Peter~J. Liu. 2020.
\newblock \href {http://arxiv.org/abs/1910.10683} {Exploring the {Limits} of
  {Transfer} {Learning} with a {Unified} {Text}-to-{Text} {Transformer}}.
\newblock Technical report.
\newblock ArXiv:1910.10683 [cs, stat] type: article.

\bibitem[{Scao et~al.(2023)Scao, Fan, Akiki, Pavlick, Ili{\'c}, Hesslow,
  Castagn{\'e}, Luccioni, Yvon, Gall{\'e}, Tow, Rush, Biderman, Webson,
  Ammanamanchi, Wang, Sagot, Muennighoff, del Moral, Ruwase, Bawden, Bekman,
  Mcmillan-Major, Beltagy, Nguyen, Saulnier, Tan, Ortiz~Suarez, Sanh, Lauren{\c
  c}on, Jernite, Launay, Mitchell, Raffel, Gokaslan, Simhi, Soroa, Aji,
  Alfassy, Rogers, Nitzav, Xu, Mou, Emezue, Klamm, Leong, van Strien, Adelani,
  Radev, Ponferrada, Levkovizh, Kim, Natan, de~Toni, Dupont, Kruszewski,
  Pistilli, Elsahar, Benyamina, Tran, Yu, Abdulmumin, Johnson, Gonzalez-Dios,
  de~la Rosa, Chim, Dodge, Zhu, Chang, Frohberg, Tobing, Bhattacharjee,
  Almubarak, Chen, Lo, von Werra, Weber, Phan, Allal, Tanguy, Dey, Mu{\~n}oz,
  Masoud, Grandury, {\v S}a{\v s}ko, Huang, Coavoux, Singh, Jiang, Vu, Jauhar,
  Ghaleb, Subramani, Kassner, Khamis, Nguyen, Espejel, de~Gibert, Villegas,
  Henderson, Colombo, Amuok, Lhoest, Harliman, Bommasani, L{\'o}pez, Ribeiro,
  Osei, Pyysalo, Nagel, Bose, Muhammad, Sharma, Longpre, Nikpoor, Silberberg,
  Pai, Zink, Torrent, Schick, Thrush, Danchev, Nikoulina, Laippala, Lepercq,
  Prabhu, Alyafeai, Talat, Raja, Heinzerling, Si, Salesky, Mielke, Lee, Sharma,
  Santilli, Chaffin, Stiegler, Datta, Szczechla, Chhablani, Wang, Pandey,
  Strobelt, Fries, Rozen, Gao, Sutawika, Bari, Al-Shaibani, Manica, Nayak,
  Teehan, Albanie, Shen, Ben-David, Bach, Kim, Bers, Fevry, Neeraj, Thakker,
  Raunak, Tang, Yong, Sun, Brody, Uri, Tojarieh, Roberts, Chung, Tae, Phang,
  Press, Li, Narayanan, Bourfoune, Casper, Rasley, Ryabinin, Mishra, Zhang,
  Shoeybi, Peyrounette, Patry, Tazi, Sanseviero, von Platen, Cornette,
  Lavall{\'e}e, Lacroix, Rajbhandari, Gandhi, Smith, Requena, Patil, Dettmers,
  Baruwa, Singh, Cheveleva, Ligozat, Subramonian, N{\'e}v{\'e}ol, Lovering,
  Garrette, Tunuguntla, Reiter, Taktasheva, Voloshina, Bogdanov, Winata,
  Schoelkopf, Kalo, Novikova, Forde, Clive, Kasai, Kawamura, Hazan, Carpuat,
  Clinciu, Kim, Cheng, Serikov, Antverg, van~der Wal, Zhang, Zhang, Gehrmann,
  Pais, Shavrina, Scialom, Yun, Limisiewicz, Rieser, Protasov, Mikhailov,
  Pruksachatkun, Belinkov, Bamberger, Kasner, Rueda, Pestana, Feizpour, Khan,
  Faranak, Santos, Hevia, Unldreaj, Aghagol, Abdollahi, Tammour, Hajihosseini,
  Behroozi, Ajibade, Saxena, Ferrandis, Contractor, Lansky, David, Kiela,
  Nguyen, Tan, Baylor, Ozoani, Mirza, Ononiwu, Rezanejad, Jones, Bhattacharya,
  Solaiman, Sedenko, Nejadgholi, Passmore, Seltzer, Sanz, Dutra, Samagaio,
  Elbadri, Mieskes, Gerchick, Akinlolu, Mckenna, Qiu, Ghauri, Burynok, Abrar,
  Rajani, Elkott, Fahmy, Samuel, An, Kromann, Hao, Alizadeh, Shubber, Wang,
  Roy, Viguier, Le, Oyebade, Le, Yang, Nguyen, Kashyap, Palasciano, Callahan,
  Shukla, Miranda-Escalada, Singh, Beilharz, Wang, Brito, Zhou, Jain, Xu,
  Fourrier, Peri{\~n}{\'a}n, Molano, Yu, Manjavacas, Barth, Fuhrimann, Altay,
  Bayrak, Burns, Vrabec, Bello, Dash, Kang, Giorgi, Golde, Posada, Sivaraman,
  Bulchandani, Liu, Shinzato, de~Bykhovetz, Takeuchi, P{\`a}mies, Castillo,
  Nezhurina, S{\"a}nger, Samwald, Cullan, Weinberg, de~Wolf, Mihaljcic, Liu,
  Freidank, Kang, Seelam, Dahlberg, Broad, Muellner, Fung, Haller,
  Chandrasekhar, Eisenberg, Martin, Canalli, Su, Su, Cahyawijaya, Garda,
  Deshmukh, Mishra, Kiblawi, Ott, Sang-Aroonsiri, Kumar, Schweter, Bharati,
  Laud, Gigant, Kainuma, Kusa, Labrak, Bajaj, Venkatraman, Xu, Xu, Xu, Tan,
  Xie, Ye, Bras, Belkada, and Wolf}]{ScaoBloom2023}
Teven~Le Scao, Angela Fan, Christopher Akiki, Ellie Pavlick, Suzana Ili{\'c},
  Daniel Hesslow, Roman Castagn{\'e}, Alexandra~Sasha Luccioni, Fran{\c c}ois
  Yvon, Matthias Gall{\'e}, Jonathan Tow, Alexander~M. Rush, Stella Biderman,
  Albert Webson, Pawan~Sasanka Ammanamanchi, Thomas Wang, Beno{\^i}t Sagot,
  Niklas Muennighoff, Albert~Villanova del Moral, Olatunji Ruwase, Rachel
  Bawden, Stas Bekman, Angelina Mcmillan-Major, Iz~Beltagy, Huu Nguyen, Lucile
  Saulnier, Samson Tan, Pedro Ortiz~Suarez, Victor Sanh, Hugo Lauren{\c c}on,
  Yacine Jernite, Julien Launay, Margaret Mitchell, Colin Raffel, Aaron
  Gokaslan, Adi Simhi, Aitor Soroa, Alham~Fikri Aji, Amit Alfassy, Anna Rogers,
  Ariel~Kreisberg Nitzav, Canwen Xu, Chenghao Mou, Chris Emezue, Christopher
  Klamm, Colin Leong, Daniel van Strien, David~Ifeoluwa Adelani, Dragomir
  Radev, Eduardo~Gonz{\'a}lez Ponferrada, Efrat Levkovizh, Ethan Kim, Eyal~Bar
  Natan, Francesco de~Toni, G{\'e}rard Dupont, Germ{\'a}n Kruszewski, Giada
  Pistilli, Hady Elsahar, Hamza Benyamina, Hieu Tran, Ian Yu, Idris Abdulmumin,
  Isaac Johnson, Itziar Gonzalez-Dios, Javier de~la Rosa, Jenny Chim, Jesse
  Dodge, Jian Zhu, Jonathan Chang, J{\"o}rg Frohberg, Joseph Tobing, Joydeep
  Bhattacharjee, Khalid Almubarak, Kimbo Chen, Kyle Lo, Leandro von Werra, Leon
  Weber, Long Phan, Loubna~Ben Allal, Ludovic Tanguy, Manan Dey, Manuel~Romero
  Mu{\~n}oz, Maraim Masoud, Mar{\'i}a Grandury, Mario {\v S}a{\v s}ko, Max
  Huang, Maximin Coavoux, Mayank Singh, Mike Tian-Jian Jiang, Minh~Chien Vu,
  Mohammad~A. Jauhar, Mustafa Ghaleb, Nishant Subramani, Nora Kassner,
  Nurulaqilla Khamis, Olivier Nguyen, Omar Espejel, Ona de~Gibert, Paulo
  Villegas, Peter Henderson, Pierre Colombo, Priscilla Amuok, Quentin Lhoest,
  Rheza Harliman, Rishi Bommasani, Roberto~Luis L{\'o}pez, Rui Ribeiro, Salomey
  Osei, Sampo Pyysalo, Sebastian Nagel, Shamik Bose, Shamsuddeen~Hassan
  Muhammad, Shanya Sharma, Shayne Longpre, Somaieh Nikpoor, Stanislav
  Silberberg, Suhas Pai, Sydney Zink, Tiago~Timponi Torrent, Timo Schick,
  Tristan Thrush, Valentin Danchev, Vassilina Nikoulina, Veronika Laippala,
  Violette Lepercq, Vrinda Prabhu, Zaid Alyafeai, Zeerak Talat, Arun Raja,
  Benjamin Heinzerling, Chenglei Si, Elizabeth Salesky, Sabrina~J. Mielke,
  Wilson~Y. Lee, Abheesht Sharma, Andrea Santilli, Antoine Chaffin, Arnaud
  Stiegler, Debajyoti Datta, Eliza Szczechla, Gunjan Chhablani, Han Wang,
  Harshit Pandey, Hendrik Strobelt, Jason~Alan Fries, Jos Rozen, Leo Gao,
  Lintang Sutawika, M~Saiful Bari, Maged~S. Al-Shaibani, Matteo Manica, Nihal
  Nayak, Ryan Teehan, Samuel Albanie, Sheng Shen, Srulik Ben-David, Stephen~H.
  Bach, Taewoon Kim, Tali Bers, Thibault Fevry, Trishala Neeraj, Urmish
  Thakker, Vikas Raunak, Xiangru Tang, Zheng-Xin Yong, Zhiqing Sun, Shaked
  Brody, Yallow Uri, Hadar Tojarieh, Adam Roberts, Hyung~Won Chung, Jaesung
  Tae, Jason Phang, Ofir Press, Conglong Li, Deepak Narayanan, Hatim Bourfoune,
  Jared Casper, Jeff Rasley, Max Ryabinin, Mayank Mishra, Minjia Zhang,
  Mohammad Shoeybi, Myriam Peyrounette, Nicolas Patry, Nouamane Tazi, Omar
  Sanseviero, Patrick von Platen, Pierre Cornette, Pierre~Fran{\c c}ois
  Lavall{\'e}e, R{\'e}mi Lacroix, Samyam Rajbhandari, Sanchit Gandhi, Shaden
  Smith, St{\'e}phane Requena, Suraj Patil, Tim Dettmers, Ahmed Baruwa,
  Amanpreet Singh, Anastasia Cheveleva, Anne-Laure Ligozat, Arjun Subramonian,
  Aur{\'e}lie N{\'e}v{\'e}ol, Charles Lovering, Dan Garrette, Deepak
  Tunuguntla, Ehud Reiter, Ekaterina Taktasheva, Ekaterina Voloshina, Eli
  Bogdanov, Genta~Indra Winata, Hailey Schoelkopf, Jan-Christoph Kalo,
  Jekaterina Novikova, Jessica~Zosa Forde, Jordan Clive, Jungo Kasai, Ken
  Kawamura, Liam Hazan, Marine Carpuat, Miruna Clinciu, Najoung Kim, Newton
  Cheng, Oleg Serikov, Omer Antverg, Oskar van~der Wal, Rui Zhang, Ruochen
  Zhang, Sebastian Gehrmann, Shani Pais, Tatiana Shavrina, Thomas Scialom, Tian
  Yun, Tomasz Limisiewicz, Verena Rieser, Vitaly Protasov, Vladislav Mikhailov,
  Yada Pruksachatkun, Yonatan Belinkov, Zachary Bamberger, Zden{\v e}k Kasner,
  Alice Rueda, Amanda Pestana, Amir Feizpour, Ammar Khan, Amy Faranak, Ana
  Santos, Anthony Hevia, Antigona Unldreaj, Arash Aghagol, Arezoo Abdollahi,
  Aycha Tammour, Azadeh Hajihosseini, Bahareh Behroozi, Benjamin Ajibade,
  Bharat Saxena, Carlos~Mu{\~n}oz Ferrandis, Danish Contractor, David Lansky,
  Davis David, Douwe Kiela, Duong~A. Nguyen, Edward Tan, Emi Baylor, Ezinwanne
  Ozoani, Fatima Mirza, Frankline Ononiwu, Habib Rezanejad, Hessie Jones,
  Indrani Bhattacharya, Irene Solaiman, Irina Sedenko, Isar Nejadgholi, Jesse
  Passmore, Josh Seltzer, Julio~Bonis Sanz, Livia Dutra, Mairon Samagaio,
  Maraim Elbadri, Margot Mieskes, Marissa Gerchick, Martha Akinlolu, Michael
  Mckenna, Mike Qiu, Muhammed Ghauri, Mykola Burynok, Nafis Abrar, Nazneen
  Rajani, Nour Elkott, Nour Fahmy, Olanrewaju Samuel, Ran An, Rasmus Kromann,
  Ryan Hao, Samira Alizadeh, Sarmad Shubber, Silas Wang, Sourav Roy, Sylvain
  Viguier, Thanh Le, Tobi Oyebade, Trieu Le, Yoyo Yang, Zach Nguyen,
  Abhinav~Ramesh Kashyap, Alfredo Palasciano, Alison Callahan, Anima Shukla,
  Antonio Miranda-Escalada, Ayush Singh, Benjamin Beilharz, Bo~Wang, Caio
  Brito, Chenxi Zhou, Chirag Jain, Chuxin Xu, Cl{\'e}mentine Fourrier,
  Daniel~Le{\'o}n Peri{\~n}{\'a}n, Daniel Molano, Dian Yu, Enrique Manjavacas,
  Fabio Barth, Florian Fuhrimann, Gabriel Altay, Giyaseddin Bayrak, Gully
  Burns, Helena~U. Vrabec, Imane Bello, Ishani Dash, Jihyun Kang, John Giorgi,
  Jonas Golde, Jose~David Posada, Karthik~Rangasai Sivaraman, Lokesh
  Bulchandani, Lu~Liu, Luisa Shinzato, Madeleine~Hahn de~Bykhovetz, Maiko
  Takeuchi, Marc P{\`a}mies, Maria~A Castillo, Marianna Nezhurina, Mario
  S{\"a}nger, Matthias Samwald, Michael Cullan, Michael Weinberg, Michiel
  de~Wolf, Mina Mihaljcic, Minna Liu, Moritz Freidank, Myungsun Kang, Natasha
  Seelam, Nathan Dahlberg, Nicholas~Michio Broad, Nikolaus Muellner, Pascale
  Fung, Patrick Haller, Ramya Chandrasekhar, Renata Eisenberg, Robert Martin,
  Rodrigo Canalli, Rosaline Su, Ruisi Su, Samuel Cahyawijaya, Samuele Garda,
  Shlok~S Deshmukh, Shubhanshu Mishra, Sid Kiblawi, Simon Ott, Sinee
  Sang-Aroonsiri, Srishti Kumar, Stefan Schweter, Sushil Bharati, Tanmay Laud,
  Th{\'e}o Gigant, Tomoya Kainuma, Wojciech Kusa, Yanis Labrak, Yash~Shailesh
  Bajaj, Yash Venkatraman, Yifan Xu, Yingxin Xu, Yu~Xu, Zhe Tan, Zhongli Xie,
  Zifan Ye, Mathilde Bras, Younes Belkada, and Thomas Wolf. 2023.
\newblock \href {https://inria.hal.science/hal-03850124} {{BLOOM: A
  176B-Parameter Open-Access Multilingual Language Model}}.
\newblock Working paper or preprint.

\bibitem[{Smith et~al.(2022)Smith, Fries, Hancock, and Bach}]{Smith2022}
Ryan Smith, Jason~A. Fries, Braden Hancock, and Stephen~H. Bach. 2022.
\newblock \href {http://arxiv.org/abs/2205.02318} {Language {Models} in the
  {Loop}: {Incorporating} {Prompting} into {Weak} {Supervision}}.
\newblock Technical report.
\newblock ArXiv:2205.02318 [cs] type: article.

\bibitem[{Sun et~al.(2023)Sun, Dong, Huang, Ma, Xia, Xue, Wang, and
  Wei}]{sun2023retentive}
Yutao Sun, Li~Dong, Shaohan Huang, Shuming Ma, Yuqing Xia, Jilong Xue, Jianyong
  Wang, and Furu Wei. 2023.
\newblock \href {http://arxiv.org/abs/2307.08621} {Retentive network: A
  successor to transformer for large language models}.

\bibitem[{Touvron et~al.(2023)Touvron, Lavril, Izacard, Martinet, Lachaux,
  Lacroix, Rozière, Goyal, Hambro, Azhar, Rodriguez, Joulin, Grave, and
  Lample}]{Touvron2023}
Hugo Touvron, Thibaut Lavril, Gautier Izacard, Xavier Martinet, Marie-Anne
  Lachaux, Timothée Lacroix, Baptiste Rozière, Naman Goyal, Eric Hambro,
  Faisal Azhar, Aurelien Rodriguez, Armand Joulin, Edouard Grave, and Guillaume
  Lample. 2023.
\newblock \href {http://arxiv.org/abs/2302.13971} {Llama: Open and efficient
  foundation language models}.

\bibitem[{Wu et~al.(2023)Wu, Waheed, Zhang, Abdul-Mageed, and Aji}]{Wu2023}
Minghao Wu, Abdul Waheed, Chiyu Zhang, Muhammad Abdul-Mageed, and Alham~Fikri
  Aji. 2023.
\newblock \href {http://arxiv.org/abs/2304.14402} {Lamini-lm: A diverse herd of
  distilled models from large-scale instructions}.
\newblock \emph{CoRR}, abs/2304.14402.

\bibitem[{Xia et~al.(2018)Xia, Zhang, Yan, Chang, and Yu}]{Xia2018}
Congying Xia, Chenwei Zhang, Xiaohui Yan, Yi~Chang, and Philip~S. Yu. 2018.
\newblock \href {https://doi.org/10.48550/arXiv.1809.00385} {Zero-shot {User}
  {Intent} {Detection} via {Capsule} {Neural} {Networks}}.
\newblock Technical report.
\newblock ArXiv:1809.00385 [cs] type: article.

\bibitem[{Zhang et~al.(2023)Zhang, Dong, Li, Zhang, Sun, Wang, Li, Hu, Zhang,
  Wu, and Wang}]{Zhang2023}
Shengyu Zhang, Linfeng Dong, Xiaoya Li, Sen Zhang, Xiaofei Sun, Shuhe Wang,
  Jiwei Li, Runyi Hu, Tianwei Zhang, Fei Wu, and Guoyin Wang. 2023.
\newblock \href {http://arxiv.org/abs/2308.10792} {Instruction {Tuning} for
  {Large} {Language} {Models}: {A} {Survey}}.
\newblock Technical report.
\newblock ArXiv:2308.10792 [cs] type: article.

\bibitem[{Zhao et~al.(2023)Zhao, Ouyang, Yu, Wu, and Li}]{Zhao2023}
Xuandong Zhao, Siqi Ouyang, Zhiguo Yu, Ming Wu, and Lei Li. 2023.
\newblock \href {https://doi.org/10.18653/v1/2023.acl-long.869} {Pre-trained
  language models can be fully zero-shot learners}.
\newblock In \emph{Proceedings of the 61st Annual Meeting of the Association
  for Computational Linguistics (Volume 1: Long Papers)}, pages 15590--15606,
  Toronto, Canada. Association for Computational Linguistics.

\end{thebibliography}

\section{Language Resource References}\label{lr:LanguageRef}
\bibliographystylelanguageresource{lrec-coling2024-natbib}
\bibliographylanguageresource{datasets}

\appendix

\section{Models}
Table~\ref{tab:decoders_only} presents decoder-only models used for classification. Table~\ref{tab:encoder_decoder} presents encoder-decoder models used for classification. Column \texttt{Model} contains the name of each model on their HuggingFace repository, column \texttt{Number of Parameters} and \texttt{Instruction-Tuned} are quite explicit.

Note that \textit{M} stands for million and \textit{B} for billion.
\label{sec:AppendixModels}
\begin{table*}[h]
    \centering
    \begin{tabular}{lrr}
        \toprule
        Model & Number of Parameters & Instruction-Tuned \\
        \midrule
        bigscience/bloom~\citep{ScaoBloom2023} & 560M, 1B1, 1B7, 3B, 7B1  & No  \\
        bigscience/bloomz~\citep{Muennighoff2023} & 560M, 1B1, 1B7, 3B, 7B1  & Yes  \\ 
        tiiuae/falcon & 7B, 40B  & Yes/No\\
        tiiuae/falcon-rw & 7B, 40B  & No\\
        MBZUAI/LaMini-Cerebras~\citep{Wu2023} & 111M, 256M, 590M, 1.3B  & Yes \\
        MBZUAI/LaMini-GPT~\citep{Wu2023} & 124M, 774M, 1.5B & Yes \\
        mosaicml/mpt & 7B 30b  & Yes/No\\
        databricks/dolly-v2 & 3b, 7B, 12b & Yes \\
        EleutherAI/pythia~\citep{Biderman2023} & 70M, 160M, 410M, 1B, 1.4B, 2.8, 6.9B, 12B  & No\\
        openlm-research/open\_llama & 3B 7B 13B & No\\
        openlm-research/open\_llama\_v2 & 3B 7B  &  No\\
        pankajmathur/orca\_dolly & 3B & Yes \\ 
        pankajmathur/orca\_alpaca & 3B & Yes \\ 
        pankajmathur/orca\_mini & 7B, 3B, 13B & Yes \\  
        pankajmathur/orca\_mini\_v2 & 7B, 13B & Yes \\  
        pankajmathur/orca\_mini\_v3 & 7B, 13B & Yes \\ 
        \bottomrule
    \end{tabular}
    \caption{Decoder Only Models}\label{tab:decoders_only}
\end{table*}

\begin{table*}[h]
    \centering
    \begin{tabular}{lrr}
        \toprule
        Model & Number of Parameters & Instruction-Tuned \\
        \midrule
        {MBZUAI/LaMini-Flan-T5}~\citep{Wu2023} & 77M, 248M, 783M & Yes  \\ 
        {T5 vanilla}~\citep{Raffel2020} & 77M, 248M, 770M, 3B, 11B & No \\ 
        {bigscience/mt0}~\citep{Muennighoff2023} & 300M, 582, 1.2B, 3.8B, 13B & Yes  \\ 
        {Bart}~\citep{Lewis2020} & 255M, 561M & No  \\ 
        \bottomrule
    \end{tabular}
    \caption{Encoder-Decoder Only Models}\label{tab:encoder_decoder}
\end{table*}

\section{Datasets}
\label{sec:AppendixDatasets}
\begin{table*}[h]
    \begin{tabular}{lrcrr}
    \toprule
    Datasets & Tasks & \#Classes & \#Test Examples & Balance ratios \\
    \midrule
    AGNews & Topic Classification & 4 & 12000 & 0.897 \\
    BBCNews & Topic Classification & 5 & 2000 & 0.742 \\
    CDR bio & Relation Classification & 2 & 4673 & 0.478 \\
    Chemprot & Chemical Relation Classification & 10 & 1607 & 0.004 \\
    ETHOS & Sentiment Classification & 2 & 998 & 0.766 \\
    financial\_phrasebank & Topic Classification & 3 & 2264 & 0.218 \\
    IMDB & Sentiment Classification & 2 & 2500 & 1.000 \\
    SemEval & Relation Classification & 9 & 600 & 0.042 \\
    SMS & Spam Classification & 2 & 500 & 0.155 \\
    Spouse & Relation Classification & 2 & 2701 & 0.088 \\
    SST2 & Sentiment Classification & 2 & 1821 & 0.997 \\
    SST5 & Sentiment Classification & 5 & 2210 & 0.441 \\
    TREC & Question Classification & 6 & 500 & 0.065 \\
    Yelp & Sentiment Classification & 2 & 3800 & 0.915 \\
    Youtube & Spam Classification & 2 & 250 & 0.894 \\
    \bottomrule
    \end{tabular}
    \caption{Datasets Descriptions}\label{tab:full_datasets}
\end{table*}

\section{Prompts \& Scoring functions}

The first is the probability of the label given the prompt, it is the most straightforward method, giving the probability of the continuation.
The second and third methods are the ratio between this probability and the probability of the label given a "tasks specific premise" (called DCPMI) and an "unconditional/not task specific premise". These methods are a reweighting of each label options according to its a priori likelihood in/out of the context of the task.
The fourth is cosine similarity, wich gives a measure of similarity between the embedding of the predicted token and the label. Th intuition behind this method is that a performant model should output a token similar to the label.

As we noticed difference in classification performances under different scoring functions but none could lead to a clear winner, could'nt juge really how well models performed. So we decided to take the mean of these scores to have a more robust evaluation of the model's performance.

\label{sec:AppendixPrompts}
% chktex-file 18
\begin{table*}[ht]
    \centering
    \caption{Prompt used}
    \resizebox{\textwidth}{!}{

        \begin{tabular}{llll}
        \toprule
        dataset & prompt & pmi\_premise & dcpmi\_premise \\
        \midrule
        sms & \texttt{Is the following message spam? Answer by yes or no.\textbackslash{n}"TEXT"} & \texttt{:} & \texttt{ The message is a spam ? } \\
        youtube & \texttt{Is the following comment spam? Answer by yes or no.\textbackslash{n}"TEXT"} & \texttt{:} & \texttt{ The comment is a spam ? } \\
        spouse & \texttt{Context: "TEXT"\textbackslash{n}\textbackslash{n}Are ENTITY2 and ENTITY1 married? Answer by yes or no.} & \texttt{:} & \texttt{Are the two entity are married? } \\
        cdr & \texttt{Context: "TEXT"\textbackslash{n}\textbackslash{n}Does ENTITY1 induce ENTITY2 ? Answer by yes or no.} & \texttt{:} & \texttt{Does the drug induce the disease? } \\
        chemprot & \texttt{Context: "TEXT"\textbackslash{n}\textbackslash{n}What is the relation between ENTITY1 and ENTITY2 ?} & \texttt{:} & \texttt{What is the relation between the two entities? } \\
        semeval & \texttt{Context: "TEXT"\textbackslash{n}\textbackslash{n}What is the relation between ENTITY1 and ENTITY2 ?} & \texttt{:} & \texttt{What is the relation between the two entities? } \\
        sst-2 & \texttt{"TEXT" has a tone that is} & \texttt{:} & \texttt{ The quote has a tone that is} \\
        sst-5 & \texttt{"TEXT" has a tone that is} & \texttt{:} & \texttt{ The quote has a tone that is} \\
        yelp & \texttt{"TEXT" has a tone that is} & \texttt{:} & \texttt{ The quote has a tone that is} \\
        imdb & \texttt{"TEXT" has a tone that is} & \texttt{:} & \texttt{ The quote has a tone that is} \\
        ethos & \texttt{"TEXT" has a tone that is} & \texttt{:} & \texttt{ The quote has a tone that is} \\
        financial\_phrasebank & \texttt{"TEXT" has a tone that is} & \texttt{:} & \texttt{ The quote has a tone that is} \\
        trec & \texttt{"TEXT" is about } & \texttt{:} & \texttt{ The topic is } \\
        agnews & \texttt{"TEXT" is about } & \texttt{:} & \texttt{ The topic is } \\
        bbcnews & \texttt{"TEXT" is about } & \texttt{:} & \texttt{ The topic is } \\
        \bottomrule
        \end{tabular}

}
\end{table*}

\end{document}